\documentclass{article}
\usepackage[rightcaption]{sidecap}
\usepackage{graphicx}
\graphicspath{ {./graphics/} }
\usepackage{fancyhdr,graphicx,amsmath,amssymb}
\usepackage[boxed]{algorithm2e}
\SetAlCapSkip{1em}
\usepackage{algpseudocode}
\usepackage{xcolor}
\usepackage{listings}

\PassOptionsToPackage{numbers, sort, square}{natbib}

\usepackage[preprint]{neurips_2019}

\usepackage[utf8]{inputenc} 
\usepackage[T1]{fontenc}    
\usepackage{hyperref}       
\usepackage{url}            
\usepackage{booktabs}       
\usepackage{amsfonts}       
\usepackage{nicefrac}       
\usepackage{microtype}      

\title{Neural Network Distiller: A Python Package For
DNN Compression Research}

\author{
  Neta Zmora\thanks{Equal contribution}\ \ \ \ Guy Jacob\footnotemark[1]\ \ \ \ Lev Zlotnik\ \ \ \ Bar Elharar\ \ \ \ Gal Novik \\
  Intel AI Lab\\
  \texttt{\{neta.zmora, guy.jacob, lev.zlotnik, bar.elharar, gal.novik\}@intel.com} \\
}

\begin{document}

\maketitle

\begin{abstract}
This paper presents the philosophy, design and feature-set of Neural Network Distiller, an open-source Python package for DNN compression research.  Distiller is a library of DNN compression algorithms implementations, with tools, tutorials and sample applications for various learning tasks. Its target users are both engineers and researchers, and the rich content is complemented by a design-for-extensibility to facilitate new research. Distiller is open-source and is available on Github at \url{https://github.com/NervanaSystems/distiller}

\end{abstract}

\section{Introduction}

Deep learning (DL) is transforming incumbent industries and creating new ones.  In many use-cases it is preferable to execute inference on network edge devices, due to privacy, bandwidth, latency, connectivity and power constraints.  However, the power, thermal, memory and compute requirements of some deep neural networks (DNN) are prohibitive for many classes of edge devices. Efforts to ameliorate this tension span multiple disciplines: Hardware \citep{DBLP:journals/corr/HanLMPPHD16}, software \citep{DLDT}, algorithm \citep{fft-conv} and DNN design and optimizations \citep{DBLP:journals/corr/HuangRSZKFFWSG016}. Neural Network Distiller is a library for DNN compression research in PyTorch \citep{paszke2017pytorch}, which caters to the latter domain: The search for ever-smaller, faster and more energy-efficient neural networks.

DNN compression is a dynamic research area with both practical and theoretical implications, making it important to industry and academia. For all practical reasons, it is important to be able to test and compare DNN compression methods under the same test conditions (datasets, pre-processing, hyper-parameters and execution environment); to experiment with mixing compression methods (e.g. quantization and pruning); and to share knowledge, results, and learnings. Distiller is our humble attempt to bring together researchers and practitioners by providing a library of algorithmic tools for DNN compression, together with tutorials, implementations of example applications on various tasks, and infrastructure code to quickly prototype new ideas. Most of the implemented algorithms relate to quantization and sparsification via pruning and regularization. Additionally, Distiller contains examples of low-rank approximation \citep{tsvd}, conditional computation \citep{Teerapittayanon_2016}, knowledge distillation \citep{hinton2015distill} and automated compression \citep{he2018amc} - and we continue adding more methods.  In this paper, we briefly describe Distiller’s motivation, design decisions and algorithms, in the hope that this work can serve others.

\section{Design}

Distiller is written in Python and is designed to be simple and extendible, accessible to experts and non-experts alike, and reusable as a library in various contexts.  An important design goal was low barriers-to-adoption by compression aficionados and researchers, for easy integration of Distiller into users’ research applications. This meant a flatter code structure, with less abstraction layers so that library code feels approachable and easy to understand. We chose PyTorch as the underlying DL framework because of its wide adoption by the research community, and opted for tight-coupling with PyTorch for simplicity and clarity at the expense of genericity.  Published works exemplify a few ways researchers are using Distiller for implementation of new compression methods \citep{zhao2019,elthakeb2019}, and for advancing their work in fields related to DNN compression \citep{shomron2019,nadeem2019,gao2019,rekhi2019}.

\begin{figure}[h]
    \centering
    \includegraphics[width=0.9\textwidth, angle=0]{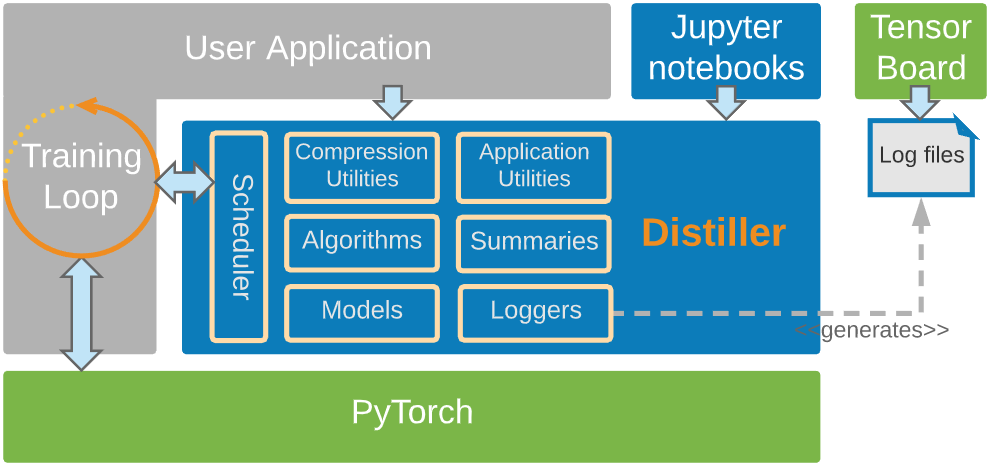}
    \caption{\textbf{Distiller high-level architecture}. Green boxes are 3rd-party components, user application code is colored gray and Distiller library code is blue.}
    \label{fig:distiller_design}
\end{figure}

Figure \ref{fig:distiller_design} depicts Distiller’s high-level architecture\footnote{More details available at \url{https://nervanasystems.github.io/distiller/design.html}}.  At the core of Distiller are implementations of the algorithmic building-blocks required for DNN compression (e.g. pruning, quantization and regularization).  These are assembled into compression methods using software components often needed for compression:

\begin{itemize}
    \item New operation layers, for example: \verb+TruncatedSVD+ as a decomposed replacement of Linear layers, and \verb+DistillerLSTM+ for fine control of quantization inside an LSTM cell
    \item Layers fusion (e.g. Batch-norm folding)
    \item Model contraction (“thinning”) and expansion
    \item Utilities such as activation statistics, graph data-dependency analysis and model summaries
\end{itemize}

For compression methods that involve training or fine-tuning models, Distiller supports various training techniques including full/partial dataset, single/multi-GPU, knowledge distillation and lottery ticket hypothesis (LTH) \citep{frankle2018lth} training.  A compression-scheduling component is responsible for interleaving training and compression, for methods such as iterative-pruning \citep{han2015learning}.

To accelerate development, Distiller also provides APIs for common utility functionalities. These include logging and plotting (e.g. per-parameter tensor sparsity); checkpointing (with compression-specific metadata) and simple experiment-artifact management; data-loading (full/partial datasets with various samplers); compression-specific command-line argument handling; and reproducible execution. We’ve integrated models from Torchvision \citep{torchvision} and Cadene’s GitHub repository \citep{cadene}, together with our own models.

We demonstrate the range and flexibility of Distiller on various example tasks and datasets: image classification,   recommendation systems (NCF \citep{he2017ncf}), NLP (e.g. GNMT \citep{wu2016gnmnt}), and object detection (e.g. Faster R-CNN \citep{ren2017faster_rcnn}).  Users can readily add models, datasets and learning tasks.

\subsection{Compression Scheduling}

Most compression methods involve some form of model training (post-training quantization being a  notable exception). These can be, for example, simple fine-tuning using a subset of the data; LTH training with some of the weights kept masked at zero; or iterative pruning following some pruning recipe. To provide the user fine-control over the interaction between the training process and the execution of compression algorithms, we built a scheduling subsystem composed of a compression-scheduler and a YAML-syntax schedule parser. The scheduler orchestrates the compression process by invoking algorithmic components, following a user-prescribed schedule. This process is demonstrated in Algorithm \ref{psuedocode_sched}.

The scheduling recipe can be defined programmatically or specified in a YAML file. Using a YAML file allows the user to quickly and easily change the compression algorithms, their configuration and scheduling, without changing code - which is especially amenable to an environment requiring rapid experimentation. The recipe defines\footnotemark the compression methods to use, and their parameters. In some cases, parameters can be overridden per-layer or per groups of layers. In addition, the recipe defines when each method should be applied - start epoch, end epoch and frequency.

\footnotetext{Examples of scheduling recipes can be found at \url{https://nervanasystems.github.io/distiller/schedule.html}}

\begin{algorithm}[h]
\DontPrintSemicolon
\SetAlgoLined
\SetNoFillComment
\label{psuedocode_sched}
\tcc{Read schedule from file}
\textcolor{blue}{\texttt{scheduler = distiller.file\_config(file\_name)}}\;
\ForEach{epoch}{
  \textcolor{blue}{\texttt{scheduler.on\_epoch\_begin(epoch)}}\;
  \ForEach{mini\_batch}{
    Get next mini-batch of data x and labels y\;
    \textcolor{blue}{\texttt{scheduler.on\_minibatch\_begin(mini\_batch)}}\;
    Clear gradients of weights\;
    Feed-forward on network M: \texttt{y' = M(x)}\;
    \tcc{Compute additive loss due to compression constraints}
    \textcolor{blue}{\texttt{compression\_loss = scheduler.before\_backward\_pass(mini\_batch)}}\;
    Compute \texttt{loss = criterion(y, y') + compression\_loss}\;
    Compute gradients\;
    \tcc{Final opportunity to alter weights before they are updated}
    \textcolor{blue}{\texttt{scheduler.before\_parameter\_optimization(mini\_batch)}}\;
    Update weights\;
    \textcolor{blue}{\texttt{scheduler.on\_minibatch\_end(mini\_batch)}}\;
   }
   Validate results using validation dataset \;
   \textcolor{blue}{\texttt{scheduler.on\_epoch\_end(epoch)}}\;
}
\caption{\textbf{Compression scheduling.} Example pseudo-code of an application invoking compression-scheduling from its training-loop. Lines in \textcolor{blue}{blue} show commands related to the scheduler, and illustrate the simplicity of adding scheduled-compression to existing PyTorch applications. Call-backs into the scheduler trigger the invocation of back-end compression algorithms according to the programmed schedule.}
\end{algorithm}

\subsection{Supporting Components}

Distiller is designed as a toolbox for the research community and is packaged with tools to share, teach and accelerate application development. Command-line utilities facilitate export to ONNX \citep{onnx} for execution in an inference-serving environment; generation of model summaries, pruning sensitivity-analysis and activation statistics.

Jupyter notebooks\footnote{\url{https://github.com/NervanaSystems/distiller/tree/master/jupyter}}, wiki tutorials\footnote{\url{https://github.com/NervanaSystems/distiller/wiki}} and rich documentation\footnote{\url{https://nervanasystems.github.io/distiller/index.html}} provide a means to share and discuss algorithmic details, explain sample applications and APIs, and explore the technical and algorithmic aspects of the compression processes and methods.

\section{Algorithms and Methods}

DNN compression is a rapidly moving field and keeping up with SoTA is challenging. Table \ref{algos-table} lists the algorithms and methods currently supported.

\begin{table}[h!]
  \caption{Compression algorithms and methods in Distiller}
  \label{algos-table}
  \centering
  \begin{tabular}{p{0.3\textwidth}p{0.6\textwidth}}
    \toprule
    Category & Methods      \\
    \midrule
    Weights Regularization      & Lp-norm, group Lasso \citep{group-lasso}, GSS \citep{torfi2018attentionbased},
                                  SSL \citep{wen2016learning} \\
    Weights Pruning             & Magnitude pruning \citep{han2015learning}, block and structure pruning, greedy pruning \citep{abbasiasl2017structural}, 
    feature-map reconstruction \citep{He_2017}, hybrid pruning \citep{hybrid},
                                 activation statistics-based
                                 \citep{hu2016network, molchanov2016pruning},
                                 Taylor expansion \citep{molchanov2016pruning},
                                 Network Surgery \citep{guo2016dynamic},
                                 automated scheduling \citep{zhu2017agp,narang2017exploring}, one-shot pruning \citep{li2016pruning}, automated-pruning \citep{he2018amc}, sensitivity analysis \citep{han2015learning}, 
                                 network thinning \citep{Leclerc2018SmallifyLN}\\
    Post-Training Quantization  & Gemmlowp-based \citep{gemmlowp,jacob2018}: dynamic / stats-based, symmetric / a-symmetric,
                                  per-tensor / per-channel, activations clipping (average min/max, ACIQ \citep{banner2019}) \\
    Quantization-Aware Training & EMA-tracked ranges \citep{jacob2018,krishnamoorthi2018}, DoReFa \citep{zhou2016dorefa}, WRPN \citep{mishra2018wrpn},
                                  PACT \citep{choi2018pact}     \\
    Conditional Computation     & Early-exit \citep{Teerapittayanon_2016}        \\
    Low-rank Decomposition     & Truncated SVD \citep{tsvd}       \\
    Others                      & Knowledge distillation \citep{hinton2015distill} \\
    \bottomrule
  \end{tabular}
\end{table}

\section{Related Work}

Support for INT8 and FP16 quantization is becoming more common, at different levels of the DL software spectrum. DL frameworks such as TensorFlow \citep{abadi2015tensorflow} and MXNet \citep{chen2015mxnet} have APIs for post-training quantization (PTQ) to INT8, with TensorFlow supporting quantization-aware training (QAT) as well. Support for built-in PTQ and QAT to INT8 was added to PyTorch very recently. NN compiler libraries such as Glow \citep{rotem2018glow} and TVM \citep{chen2018tvm}, and platform-specific solutions such as Intel OpenVino \citep{openvino} and NVIDIA TensorRT \citep{tensorrt} provide PTQ of FP32 models to INT8/FP16. Support for pruning is far less prevalent. TensorFlow provides a model pruning API which is based on a specific magnitude-pruning algorithm \citep{zhu2017agp}. PocketFlow \citep{wu2018pocketflow} is a framework for automatic model compression and acceleration which has some overlapping features with Distiller, with both pruning and quantization capabilities.

Compared to the above, Distiller provides a larger collection and mix of compression methods implemented out-of-the-box. Distiller offers flexibility and control over the compression process via its scheduling mechanism and YAML configuration. This makes it easy to experiment with, for example, mixed-precision quantization at the layer level, or with different pruning settings for different layers at different stages of training. Finally, beyond ready-to-use algorithms, Distiller implements the infrastructure and exposes the basic building blocks which can be used to develop new ones.

A disadvantage of Distiller's current implementation of PTQ is that it is simulating INT8 using FP32 operations. In addition, it is lacking the capability to export quantized model to ONNX and/or Glow. This makes Distiller a great vehicle to evaluate the effects of quantization on a model, but makes it hard to realize the gains from quantization on actual hardware. With the upcoming release of native quantization capabilities in PyTorch, we hope to rectify this.

\section{Conclusion and Future Work}

We presented Neural Network Distiller, a DNN-compression research library built for the community at large, as part of Intel AI Lab’s contribution to the DL community.  Distiller packages together compression algorithms and methods, a supporting training framework and examples of model compression for various learning tasks. 
Distiller is a work-in-progress and there is room for improvement.  Compression automation is an emerging research field with an exciting promise that we intend to follow very closely. 

Progress in DNN compression research can provide the tailwind necessary to help bring DL innovation to more industries and application domains, to make people’s lives easier, healthier, and more productive.  Distiller is an attempt to build a community of researchers, scientists and engineers vested in this future vision. 

\subsubsection*{Acknowledgements}
We wish to thank Yury Nahshan for the implementation of ACIQ; Haim Barad for the implementation of Early-exit; and the Distiller community on GitHub for bug fixes, small features and challenging questions and observations.

\small{
  \bibliography{distiller}
  \bibliographystyle{unsrtnat}
}

\end{document}